\documentclass[letterpaper, 10 pt, conference]{ieeeconf}  

\IEEEoverridecommandlockouts                              

\usepackage{epsfig} 

\usepackage{subfig}
\usepackage{amssymb}
\usepackage{cite}

\newcommand{\argmin}{\mathop{\rm arg~min}\limits}

\title{\LARGE \bf
Multiple Human Tracking using Multi-Cues \\
including Primitive Action Features
}

\author{Hitoshi Nishimura$^{1,2,a}$ Kazuyuki Tasaka$^{2,b}$ Yasutomo Kawanishi$^{1,2,c}$ Hiroshi Murase$^{1,2,d}$
\thanks{$^{1}$KDDI Research, Inc.}%
\thanks{$^{2}$Graduate School of Informatics, Nagoya University}%
\thanks{$^{a}$\tt\small ht-nishimura@kddi-research.jp}
\thanks{$^{b}$\tt\small ka-tasaka@kddi-research.jp}
\thanks{$^{c}$\tt\small kawanishi@is.nagoya-u.ac.jp}
\thanks{$^{d}$\tt\small murase@nagoya-u.jp}
}

\begin{document}
\maketitle
\thispagestyle{empty}
\pagestyle{empty}

\begin{abstract}
In this paper, we propose a Multiple Human Tracking method using multi-cues including Primitive Action Features (MHT-PAF).
MHT-PAF can perform the accurate human tracking in dynamic aerial videos captured by a drone.
PAF employs a global context, rich information by multi-label actions, and a middle level feature.
The accurate human tracking result using PAF helps multi-frame-based action recognition.
In the experiments, we verified the effectiveness of the proposed method using the Okutama-Action dataset.
Our code is available online\footnote{https://github.com/hitottiez/mht-paf}.
\end{abstract}

\section{Introduction}
Multiple human tracking is a fundamental technique and widely used in various fields such as robotics, surveillance, and marketing.
The task of multiple human tracking is that multiple humans continue to be detected, maintaining their identities (ID) base on time-series images~\cite{collins2000system}.
Most state-of-the-art tracking methods~\cite{breitenstein2010online,Wojke2017simple,zhang2008global,berclaz2011multiple,milan2013continuous} are based on a tracking-by-detection approach owing to the recent improvement of in accuracy of human detection.
The tracking-by-detection approach considers multiple human tracking as data association \cite{zhang2008global}.
The data association matches detection results between consecutive frames with an association metric.

In aerial images captured by a drone, human tracking accuracy is not so accurate.
i) significant change of human's size and aspect ratio and ii) abrupt camera movement cause false positives and ID switches.
ID switch means the target human ID changes to another ID.
Although human tracking methods~\cite{breitenstein2010online,Wojke2017simple,zhang2008global,berclaz2011multiple,milan2013continuous} utilize a human appearance feature, position, or both of them as cues, the false positive and ID switch frequently occur.

\begin{figure}[t]
 \begin{center}
\subfloat[Previous methods]{
 \includegraphics[width=0.352\linewidth]{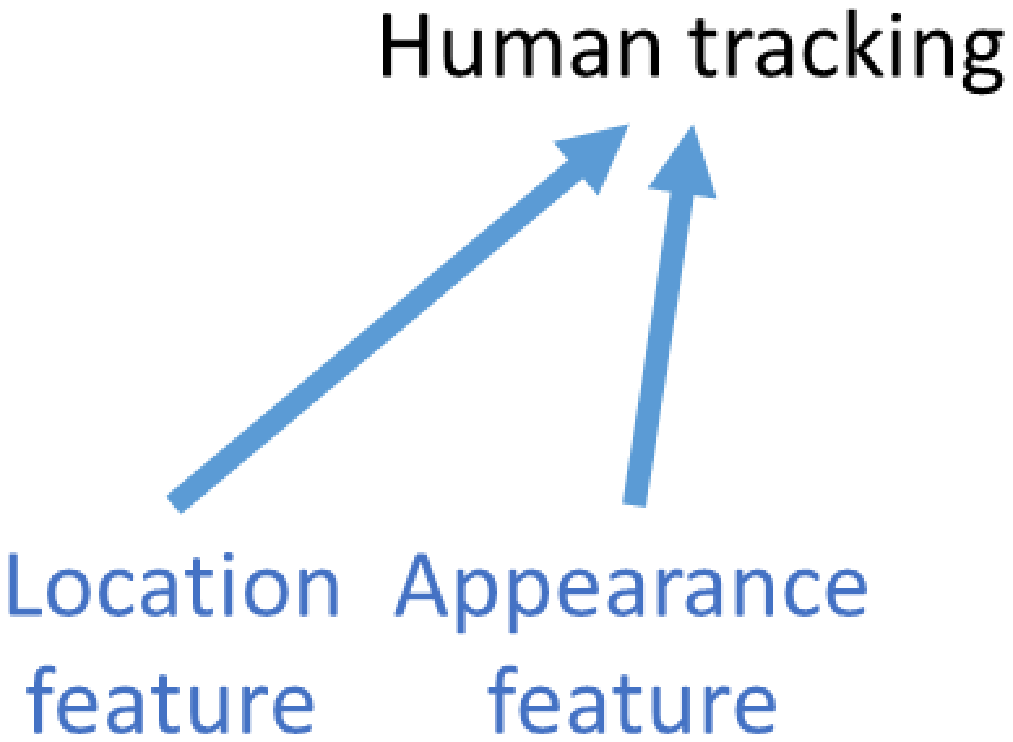}
 \label{fig:idea1}
 }
\hspace{-5mm}
\subfloat[Proposed MHT-PAF]{
 \includegraphics[width=0.622\linewidth]{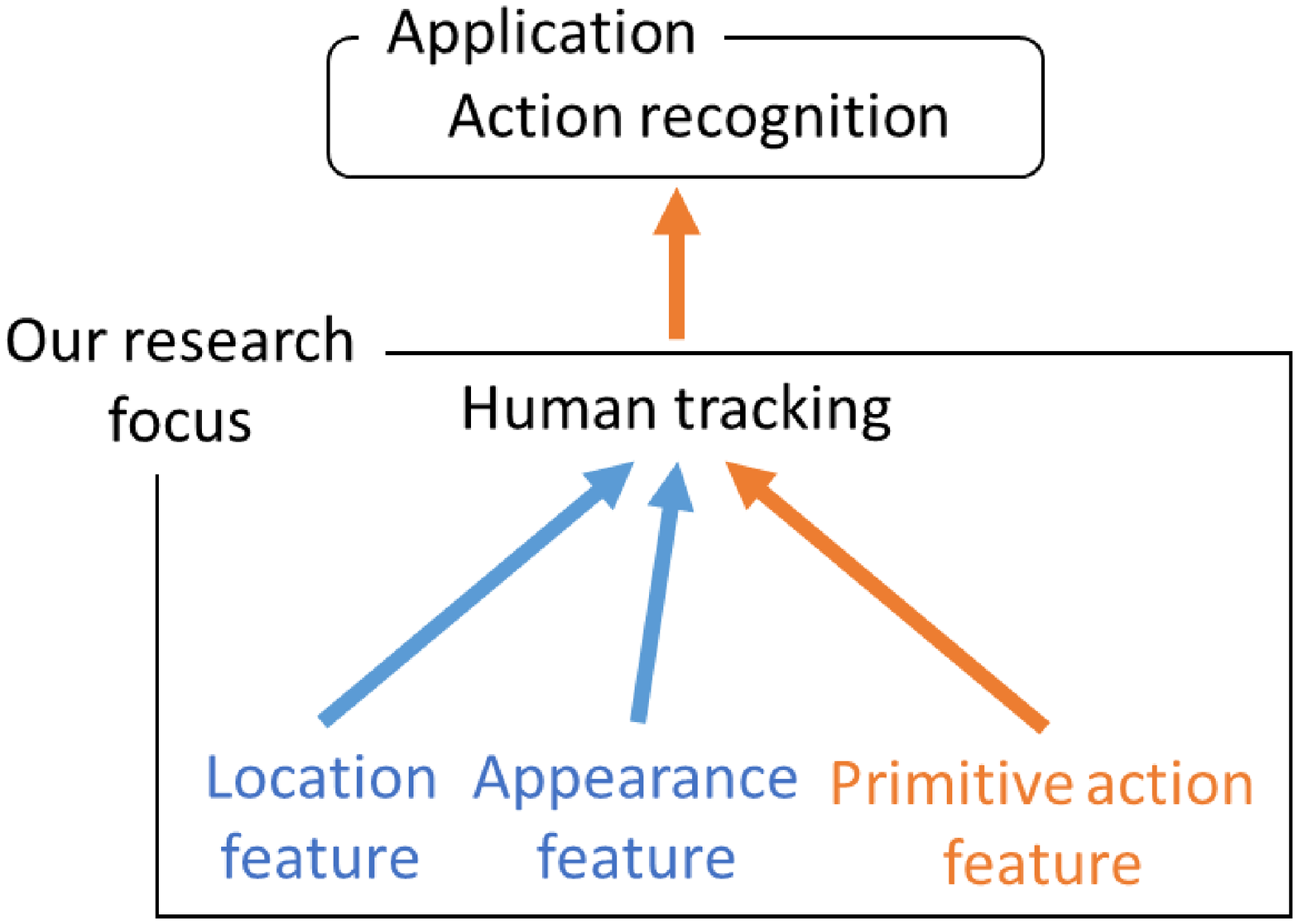}
 \label{fig:idea2}
 }
 \end{center}
 \caption{Idea of the proposed method.}
 \label{fig:idea}
\end{figure}

In this paper, we propose a Multiple Human Tracking method using multiple cues, including Primitive Action Features (MHT-PAF).
Our idea is that an action cue is effective for each human's tracking because a human action does not change frequently at the frame level.
Fig.\ \ref{fig:idea} shows the idea of the proposed method.
Unlike previous methods (Fig.\ \ref{fig:idea}\subref{fig:idea1}), the proposed method employs the action feature for human tracking (Fig.\ \ref{fig:idea}\subref{fig:idea2}).
For data association, the action feature needs to be extracted for each frame.

However, the frame-level feature works poorly because a human action occurs across multiple frames.
Therefore, we designed the primitive action feature (PAF) in terms of the following three points:
\begin{enumerate}
\item PAF employs a spatial context with a global cropped image in order to capture actions including interactions with human and object.
\item PAF is based on multi-label actions in order to extract rich information on human action.
\item PAF employs not a final action recognition label but a middle level feature because a human action includes some ambiguity at the frame level.
\end{enumerate}
Accurate human tracking using PAF can be applied to the multi-frame-based action recognition described in Fig.\ \ref{fig:idea}\subref{fig:idea2}.

Our main contributions are as follows:
\begin{itemize}
\item We propose a multiple human tracking method using multi-cues that include primitive action features  (MHT-PAF).
\item We design PAF that employs a global context, rich information by multi-labeling of actions, and a middle level feature.
\item We verify the effectiveness of the proposed method and make the code available on the web\footnotemark[1].
\end{itemize}

\section{Related Work}
Human Tracking:
In human tracking, a human continue to be detected, maintaining the ID.
The human tracking methods are classified into online and offline methods.
While online methods estimate the human ID in a serial fashion, offline methods estimate it after all data is stored.
Online:
Breitenstein {\sl et al.} introduced human tracking in a particle filtering framework~\cite{breitenstein2010online}.
Wojke {\sl et al.} proposed DeepSORT, which performs human tracking using human features and positions~\cite{Wojke2017simple}.
Offline:
Zhang {\sl et al.} proposed MCF, which solves human tracking as a minimum-cost flow problem~\cite{zhang2008global}. 
Berclaz {\sl et al.} reformulated human tracking as a constrained flow optimization in a convex problem~\cite{berclaz2011multiple}. 
Milan {\sl et al.} proposed a human tracking method that is solved by continuous energy minimization~\cite{milan2013continuous}.
In these previous methods, the accuracy of human tracking is not so accurate because the methods utilize only human appearance features, positions, or both of them as cues.

Action Detection:
In action detection, spatio-temporal action positions and action classes are estimated.
Many action detection methods have been proposed~\cite{gkioxari2015finding,lin2017single,donahue2015long,shou2016temporal,hou2017tube,kalogeiton2017action,singh2017online}.
Action detection is classified into three categories.
(1) Spatial action detection:
Gkioxari {\sl et al.} introduced action tubes that operate with region proposals, CNN features, and SVMs~\cite{gkioxari2015finding}.
Lin {\sl et al.} proposed SSAD, which is an end-to-end neural network~\cite{lin2017single}.
(2) Temporal action detection:
LRCN performs temporal action detection using LSTM~\cite{donahue2015long}.
Shou {\sl et al.} introduced Mulit-stage CNN, which employs 3D CNNs for temporal action detection~\cite{shou2016temporal}.
(3) Spatio-Temporal action detection:
Hou {\sl et al.} proposed T-CNN, which is a unified deep neural network that detects actions based on 3D convolution features~\cite{hou2017tube}.
Kalogeiton {\sl et al.} proposed an ACT detector which is also a unified deep neural network, and based on stacking of single-frame features~\cite{kalogeiton2017action}.
Singh {\sl et al.} presented ROAD, which performs spatio-temporal action detection online~\cite{singh2017online}.
All these methods utilize action information as a cue, but not a specific human appearance feature that captures human ID as cues.

Action Recognition:
For action recognition, an action class is estimated, given a spatio-temporal action position.
Many action recognition methods have been proposed~\cite{simonyan2014two,tran2015learning,donahue2015long,wang2016temporal,carreira2017quo}. 
Simonyan {\sl et al.} introduced a two-stream network using RGB and flow images~\cite{simonyan2014two}.
The proposed method is based on a two-stream network because of its simplicity.
Wang {\sl et al.} proposed TSN, which divides an image into several segments in a temporal domain~\cite{wang2016temporal}.
Donahue {\sl et al.} introduced LRCN, which performs long-term action recognition using LSTM~\cite{donahue2015long}.
Tran {\sl et al.} proposed C3D, which extracts a feature by 3D convolution~\cite{tran2015learning}.
Carreira {\sl et al.} proposed I3D, which uses a 3D convolution, parameters of which are based on 2D convolution~\cite{carreira2017quo}.

\section{Proposed Method (MHT-PAF)}
\begin{figure}[t]
 \begin{center}
 \includegraphics[width=0.99\linewidth]{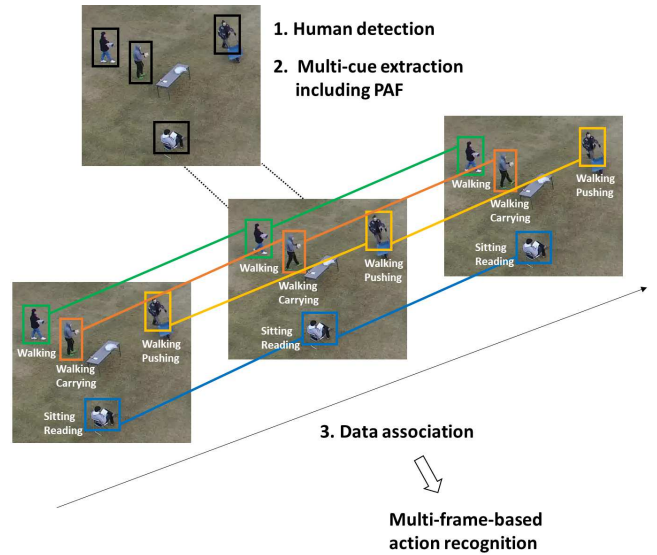}
 \end{center}
 \caption{Pipeline of the proposed method.}
 \label{fig:proposed}
\end{figure}

In order to prevent false positives and ID switches, we introduce the primitive action feature (PAF) for human tracking.
First of all, we explain the problem formulation (Section\ \ref{formulation}).
Fig.\ \ref{fig:proposed} shows the pipeline of the proposed method.
Multi-cues including PAF are extracted (Section\ \ref{multi_cues}).
After this procedure is completed for all frames, data association is performed (Section\ \ref{data_association}).
The data association results can be applied to the multi-frame-based action recognition (Section\ \ref{multi_action}).

\subsection{Problem Formulation} \label{formulation}
Let $Y=\{\mathbf{y}_{i}\}$ be a set of human observations, each of which is a human detection result.
The $i$-th observation is defined as $\mathbf{y}_i = (t_i, \mathbf{x}^{{\rm loc}}_i, \mathbf{x}^{{\rm app}}_i, \mathbf{x}^{{\rm paf}}_i)$.
$t$ denotes a time step.
$\mathbf{x}^{{\rm loc}} = (x, y, w, h)$ denotes the bounding box of a human.
while $x$ and $y$ are the x and y coordinates of the upper left corner of a rectangle,
and $w$ and $h$ are its width and height.
$\mathbf{x}^{{\rm app}}$ and $\mathbf{x}^{{\rm paf}}$ denote an appearance feature and a primitive action feature, respectively.
Let $Y_k=(\mathbf{y}_{k_1}, \mathbf{y}_{k_2}, \cdots, \mathbf{y}_{k_{l_k}})$ be the $k$-th human trajectory.
Human tracking estimates all trajectories $\Omega=\{Y_k\}$, given time series images. 

\subsection{Multi-cue Extraction} \label{multi_cues}
For each $\mathbf{y}_{i}$, three types of cues are extracted, a location feature ($\mathbf{x}^{{\rm loc}}_i$), an appearance feature ($\mathbf{x}^{{\rm app}}_i$), and a primitive action feature ($\mathbf{x}^{{\rm paf}}_i$).

\subsubsection{Location Feature} \label{location_feature}
Each bounding box $\mathbf{x}^{{\rm loc}}_i$ is estimated by SSD~\cite{liu2016ssd}.
The backbone model of the SSD is VGG16~\cite{simonyan2014very}.
For the input, we use a 4K image in order to capture human actions in detail.

\subsubsection{Appearance Feature} \label{appearance_feature}
The appearance feature $\mathbf{x}^{{\rm app}}_i$ is defined in a feature space where the distance between two features is small when their features indicate the same human.
It is extracted by a Siamese network that has $2$ inputs and $1$ output~\cite{wojke2018deep}.
Each backbone model of the Siamese network is WideResNet~\cite{zagoruyko2016wide}.
In the training phase, while the same human pair is annotated with a ``1'', a different human pair is annotated with a ``0''.
In the inference phase, one of the two backbone models is used for extracting the appearance feature.

\subsubsection{Primitive Action Feature} \label{primitive_feature}
\begin{figure}[t]
 \begin{center}
 \includegraphics[width=0.99\linewidth]{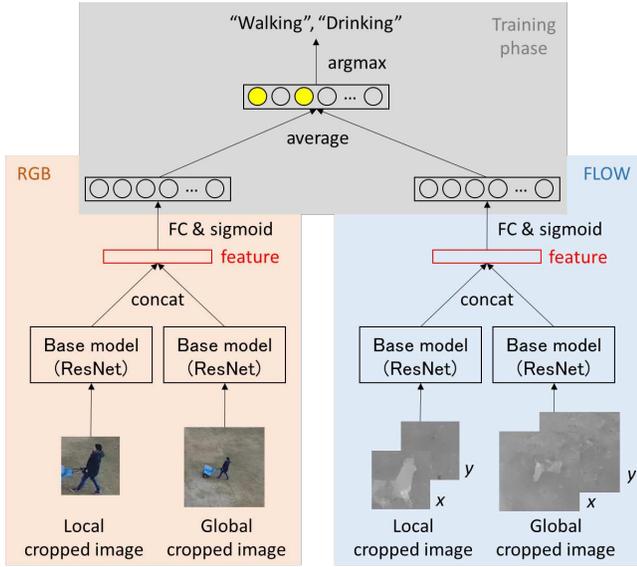}
 \end{center}
 \caption{Primitive action feature extraction model.}
 \label{fig:primitive_feature}
\end{figure}

In this section, we describe the primitive action feature (PAF), which is the key point of the proposed method.
A human region is cropped corresponding to $\mathbf{x}^{{\rm loc}}_i$.
For each cropped image, a primitive action feature $\mathbf{x}^{{\rm paf}}_i$ is extracted.
Fig.\  \ref{fig:primitive_feature} shows the PAF extraction model.
PAF employs a global context, rich information by multi-label actions, and a middle level feature.
The network is a four-stream neural network.

The network is base on two-stream network~\cite{simonyan2014two,wang2016temporal} which has two modalities, spatial and temporal modalities.
While the spatial network utilizes a RGB image, the temporal network utilizes an optical flow image.
For optical flow calculation, we used TV-L1 optical flow~\cite{zach2007duality} which is a fast and accurate method.
The optical flow is calculated in x and y coordinates separately.
The backbone model of each stream is ResNet101~\cite{he2016deep}.

For each modality, two types of images are input to the network, local and global cropped images.
The local cropped image is obtained from a bounding box which is squared from $\mathbf{x}^{{\rm loc}}_i$ to fit the long side.
The global cropped image is the expanded image from the local cropped image, regarding $\mathbf{x}^{{\rm loc}}_i$ as the center.
The expansion ratio is set as a predefined parameter $\mu$.
The global cropped image introduces the spatial context such as objects and other humans.

Since the output of the network are multi-label actions, the feature extracts rich information on human action.
The loss function is a binary cross entropy loss for each class.

The primitive action feature (RGB) is directly extracted from the layer just before the fully connected layer in the RGB network.
As well as the primitive action feature (RGB) itself, the primitive action feature (FLOW) is extracted from the FLOW network.
The primitive action feature $\mathbf{x}^{{\rm paf}}_i$ is obtained by concatenating  the RGB and FLOW features.

\subsection{Data Association} \label{data_association}
Data association is performed based on the multi-cues described in Section\ \ref{multi_cues}.
Data association is regarded as a minimum-cost flow problem~\cite{zhang2008global}.
We define four types of costs, $c_{\rm obsv}(i)$, $c_{\rm tran}(i, j)$, $c_{\rm entr}(i)$, and $c_{\rm exit}(i)$.

The first, $c_{\rm obsv}(i)$, is the observation cost of the $i$-th observation
and is based on the logistic function as follows:
\begin{equation}
p = \frac{1}{1 + \exp(b + \alpha + \beta\cdot c_{\rm det}(i))},
\end{equation}
\begin{equation}
c_{\rm obsv}(i) = -\log \frac{p}{1-p},
\end{equation}
where $b$ denotes a predefined bias, $c_{\rm det}$ denotes the score of the human detection, and $\alpha$ and $\beta$ denote parameters of the logistic function.
In the training phase, $\alpha$ and $\beta$ are estimated by the Fisher scoring algorithm.

$c_{\rm tran}(i, j)$ is the transition cost between the $i$-th observation and the $j$-th observation
and is based on the nonlinear function $g$ as follows:
\begin{equation}
q = g(c_{\rm iou}(i, j), c_{\rm app}(i, j), c_{\rm paf}(i, j)),
\end{equation}
\begin{equation}
c_{\rm tran}(i, j) = -\log({\rm sigmoid}(q)),
\end{equation}
where $c_{\rm iou}$, $c_{\rm app}$, and $c_{\rm paf}$ respectively denote
an IoU (Intersection over Union) score, a cosine distance of an appearance feature, and a cosine distance of a primitive action feature.
$g$ is represented by multiple decision trees.
In the training phase, the parameters of $g$ are estimated by a gradient boosting algorithm~\cite{friedman2001greedy}.

$c_{\rm entr}(i)$ is the entry cost of the $i$-th observation, and $c_{\rm exit}(i)$ is the exit cost of the $i$-th observation.

Human tracking is performed by estimating a set of indicator variables $F$ as follows:
\begin{equation}
F=\{f_{\rm entr}(i), f_{\rm obsv}(i), f_{\rm tran}(i, j), f_{\rm exit}(i)\ |\ \forall i, j\},
\end{equation}
where $f_{\rm entr}(i), f_{\rm obsv}(i), f_{\rm tran}(i, j), f_{\rm exit}(i) \in \{0,1\}$~\cite{zhang2008global}.
$F$ is estimated by minimizing the following objective function with non-overlap constraints~\cite{zhang2008global}:
\begin{eqnarray}
F^* = \argmin_{F} ⁡\sum_i c_{\rm entr}(i)f_{\rm entr}(i) + \sum_i c_{\rm obsv}(i)f_{\rm obsv}(i) \nonumber \\
+ ⁡\sum_i \sum_j c_{\rm tran}(i, j)f_{\rm tran}(i, j) + ⁡\sum_i c_{\rm exit}(i)f_{\rm exit}(i).
\end{eqnarray}
The objective function is minimized by the scaling push-relabel method.
In the online solution, the objective function is solved by the Hungarian algorithm for each frame.

\begin{table*}[t]
\begin{center}
\caption{Performance of human tracking.}
\label{tb:data_association}
\scalebox{1.00}{
\begin{tabular}{l||r|r|r|r|r}
\hline
 & \multicolumn{1}{c|}{Recall ($\%$) $\uparrow$} & \multicolumn{1}{c|}{Precision ($\%$) $\uparrow$} & \multicolumn{1}{c|}{IDs $\downarrow$} & \multicolumn{1}{c|}{FM $\downarrow$} & \multicolumn{1}{c}{MOTA $\uparrow$} \\
\hline
\hline
DeepSORT \cite{Wojke2017simple} & $38.04$ & $68.79$ & $1034$ & $2481$ & $20.33$ \\
\hline
MCF \cite{zhang2008global} & ${\bf 39.62}$ & $67.64$ & $496$ & $1875$ & $20.71$ \\
\hline
MHT-PAF (late) & $39.32$ & $67.51$ & $504$ & $1902$ & $20.45$ \\
\hline
MHT-PAF & $39.07$ & ${\bf 70.15}$ & ${\bf 386}$ & ${\bf 1833}$ & ${\bf 22.94}$ \\
\hline
\end{tabular}
}
\end{center}
\end{table*}

\begin{table*}[t]
\begin{center}
\caption{Average Precision (AP) of multi-frame-based action detection ($\%$).}
\label{tb:multi}
\scalebox{0.850}{
\begin{tabular}{l|r|r|r|r|r|r|r|r|r|r|r|r|r}
\hline
& \multicolumn{2}{c|}{Human to human interactions} & \multicolumn{5}{|c|}{Human to object interactions} & \multicolumn{5}{|c|}{No-interaction} & Mean \\
\cline{2-13}
& \multicolumn{1}{c|}{handshaking} & \multicolumn{1}{c|}{hugging} & \multicolumn{1}{c|}{reading} & \multicolumn{1}{c|}{drinking} & \multicolumn{1}{c|}{pushing/pulling} & \multicolumn{1}{c|}{carrying} & \multicolumn{1}{c|}{calling} & \multicolumn{1}{c|}{running} & \multicolumn{1}{c|}{walking} & \multicolumn{1}{c|}{lying} & \multicolumn{1}{c|}{sitting} & \multicolumn{1}{c|}{standing} & (mAP) \\
\hline
\hline
DeepSORT \cite{Wojke2017simple} & ${\bf 0.03}$ & ${\bf 0.50}$ & ${\bf 8.26}$ & $0$ & $18.02$ & $16.97$ & $3.16$ & $23.65$ & $31.36$ & $0$ & ${\bf 23.67}$ & $16.11$ & $11.81$ \\
\hline
MCF \cite{zhang2008global} & $0$ & $0.48$ & $5.60$ & $0$ & $18.54$ & ${\bf 18.41}$ & $3.07$ & $29.34$ & $30.27$ & $0$ & $21.16$ & ${\bf 16.78}$ & $11.97$ \\
\hline
MHT-PAF (late) & $0$ & $0.35$ & $5.59$ & $0$ & $17.86$ & $16.78$ & $3.07$ & $29.37$ & $30.24$ & ${\bf 0.48}$ & $21.11$ & $16.01$ & $11.74$ \\
\hline
MHT-PAF & $0$ & $0.34$ & $7.40$ & $0$ & ${\bf 19.69}$ & $17.78$ & ${\bf 3.55}$ & ${\bf 30.78}$ & ${\bf 31.65}$ & $0$ & $22.92$ & $16.42$ & ${\bf 12.54}$ \\
\hline
\end{tabular}
}
\end{center}
\end{table*}

\subsection{Multi-frame-based Action Recognition} \label{multi_action}
\begin{figure}[t]
 \begin{center}
 \includegraphics[width=0.85\linewidth]{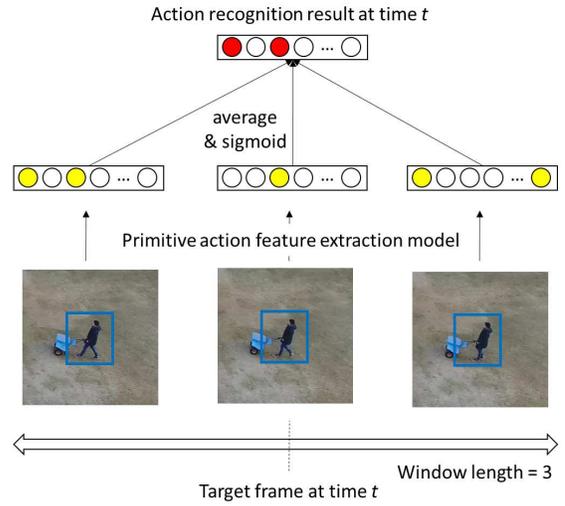}
 \end{center}
 \caption{Multi-frame-based action recognition.}
 \label{fig:multi}
\end{figure}

The human tracking result described in Section\ \ref{data_association} is applied to the multi-frame-based action recognition.
For each $\mathbf{y}_{i}$, $a_i$ which denotes an action class is estimated.
Fig.\ \ref{fig:multi} shows multi-frame-based action recognition, which is based on a sliding window.
At time $t$, the average action recognition score within the window is calculated.
Each action recognition score is directly extracted from the last class layer of PAF extraction model.
Then, the action recognition result is estimated.
The window length $\lambda$ is a predefined parameter.
When the action recognition score is lower than predefined threshold $\epsilon$, the action is determined to be ``Unknown''.

\section{Experiments}
We conducted experiments of the human tracking in order to verify the effectiveness of the proposed human tracking method and its usefulness for the multi-frame-base action recognition.

\subsection{Dataset}
We used an Okutama-Action dataset~\cite{barekatain2017okutama}, which is an aerial view of a concurrent human action detection dataset.
The dataset is very challenging because it includes significant changes in human's size and aspect ratio, and abrupt camera movement, dynamic transition of multi-label actions.
The dataset contains $43$ videos and was split into the training ($33$ videos) and test data ($10$ videos).
The videos are captured in $30$ FPS, and $77,365$ images in the dataset in total.
Two drones captured $9$ participants from a distance of $10$-$45$ meters and camera angles of $45$-$90$ degrees.
The resolution of images is 4K ($3,840 \times 2,160$).
Each bounding box have one or more action labels.
Twelve action labels are divided into three categories:
human-to-human interactions (handshaking, hugging), human-to-object interactions (reading, drinking, pushing/pulling, carrying, calling), and no-interaction (running, walking, lying, sitting, standing).
Multiple actions almost always consist of one no-interaction action and one action from the other two categories.

\begin{table*}[t]
\begin{center}
\caption{Accuracy of action recognition, given ground truth of human tracking ($\%$).}
\label{tb:action}
\scalebox{0.790}{
\begin{tabular}{l|r|r|r|r|r|r|r|r|r|r|r|r|r}
\hline
& \multicolumn{2}{c|}{Human to human interactions} & \multicolumn{5}{|c|}{Human to object interactions} & \multicolumn{5}{|c|}{No-interaction} & Average \\
\cline{2-13}
& \multicolumn{1}{c|}{handshaking} & \multicolumn{1}{c|}{hugging} & \multicolumn{1}{c|}{reading} & \multicolumn{1}{c|}{drinking} & \multicolumn{1}{c|}{pushing/pulling} & \multicolumn{1}{c|}{carrying} & \multicolumn{1}{c|}{calling} & \multicolumn{1}{c|}{running} & \multicolumn{1}{c|}{walking} & \multicolumn{1}{c|}{lying} & \multicolumn{1}{c|}{sitting} & \multicolumn{1}{c|}{standing} & \\
\hline
\hline
Single frame (local) & $7.78$ & $21.47$ & $57.26$ & $0$ & $55.95$ & $53.57$ & $15.08$ & $43.63$ & $79.97$ & $24.19$ & $76.31$ & $79.37$ & $42.88$ \\
\hline
Single frame (local+global) & $17.11$ & $24.60$ & $61.27$ & $0$ & $64.31$ & $74.13$ & ${\bf 17.82}$ & $41.18$ & $85.17$ & $14.33$ & $75.41$ & $75.89$ & $45.94$ \\
\hline
Multi frames (local) & $9.64$ & $17.68$ & $56.75$ & $0$ & $60.88$ & $56.94$ & $13.97$ & ${\bf 46.72}$ & $87.19$ & ${\bf 26.56}$ & ${\bf 81.94}$ & ${\bf 82.83}$ & $45.09$ \\
\hline
Multi frames (local+global) & ${\bf 18.07}$ & ${\bf 24.81}$ & ${\bf 61.37}$ & $0$ & ${\bf 68.28}$ & ${\bf 78.20}$ & $16.76$ & $43.55$ & ${\bf 90.73}$ & $15.40$ & $78.04$ & $78.40$ & ${\bf 47.80}$ \\
\hline
\end{tabular}
}
\end{center}
\end{table*}

\subsection{Experimental Setting}
The human detection model (SSD) was trained using the Okutama-Action dataset.
It was trained for $6,000$ iterations with a learning rate of $10^{-4}$.
The input size of SSD was $512 \times 512$.
We used the same human detection results for the previous methods and the proposed method.
The appearance feature extraction model (WideResNet) was trained using the MARS dataset~\cite{zheng2016mars}.
The primitive action feature (PAF) extraction model was trained using the Okutama-Action dataset.
It was trained for $5,000$ iterations with a learning rate of $10^{-4}$.
The dropout ratio was set to $0.7$.
In the data augmentation, random cropping and horizontal/vertical cropping were employed.
$\mu=3$, $\epsilon=0.4$, $\lambda=15$.
The dimension of PAF was $4096$ (RGB: $2048$$ + $FLOW: $2048$).
The observation cost model and the transition cost model were trained using Okutama-Action dataset.
For data association parameters, we empirically set $c_{\rm entr}(i)=10$, $c_{\rm exit}(i)=10$, $b=-2$.

\subsection{Evaluation of Proposed Human Tracking}
We evaluated the human tracking (Estimating $\mathbf{x}^{{\rm loc}}$ and $f$).
For the evaluation metric, we used Multiple Object Tracking Accuracy (MOTA).
MOTA is a widely used and comprehensive metric using the following combination:
\begin{equation}
MOTA = 1 - (FN + IDs + FP)\ /\ DET,
\end{equation}
where FN, IDs, FP, and DET respectively denote the total number of false negatives, ID switches, false positives, and detections.
The MOTA score ranges from $-\infty$ to $100$.
More details about these metrics are described in paper~\cite{bernardin2008evaluating}.
The IoU threshold between ground truth and the estimated bounding box was set to $0.5$.

Table\ \ref{tb:data_association} shows the performance of human tracking.
The MOTA in the case of MHT-PAF is higher than that of MCF~\cite{zhang2008global}, which does not utilize the action feature (MCF: $20.71$ vs. MHT-PAF: $22.94$).
While keeping the recall almost same (MCF: $39.62$ vs. MHT-PAF: $39.07$), the precision improved $2.51$ (MCF: $67.64$ vs. MHT-PAF: $70.15$).
The number of ID switches decreased $110$ (MCF: $496$ vs. MHT-PAF: $386$), and the fragmentation decreased $42$ (MCF: $1875$ vs. MHT-PAF: $1833$).

MHT-PAF (late) employs a late fusion which concatenates RGB and FLOW of the last class layer for the action feature.
The MOTA in the case of MHT-PAF is higher than MHT-PAF (late) (MCF-PAF (late): $20.45$ vs. MHT-PAF: $22.94$).
Also, the MOTA in the case of MHT-PAF is higher than that of DeepSORT~\cite{Wojke2017simple} (DeepSORT: $20.33$ vs. MHT-PAF: $22.94$).

\subsection{Application for Multi-Frame-based Action Recognition}
We evaluated the multi-frame-based action recognition.
For the evaluation metric, we used mean Average Precision (mAP).
The mAP is used for the action detection task, which estimates $\mathbf{x}^{{\rm loc}}$ and $a$.
The IoU threshold between ground truth and estimated bounding box was set to $0.5$.
%
Table\ \ref{tb:multi} shows the result of multi-frame-based action detection.
In DeepSORT and MCF, PAF was not utilized.
The mAP in the case of MHT-PAF is higher than that of MCF (MCF: $11.97$ vs. MHT-PAF: $12.54$).
This is due to the improvement in accuracy of human tracking using PAF.

\subsection{Discussion}
We evaluated the accuracy of action recognition (Estimating $a$), given the ground truth of human tracking.
The purpose was to discuss the global cropped image and single/multi-frame-based action recognition.
The evaluation was performed at frame level.
Table\ \ref{tb:action} shows the accuracy of action recognition.

Global Cropped Image:
Let us compare the local cropped image to local+global cropped image in the single-frame-based action recognition.
The accuracy in the case of local+global cropped image is higher than that of local cropped image
(local: $42.88$ vs. local+global: $45.94$).
For human-to-human interactions and human-to-object interactions, the global cropped image is effective.
These interactions need a global context such as humans or objects for recognition.
On the other hands, for no-interaction, the local cropped image is effective.
No-interaction needs only human motions for recognition.
The average accuracy is the highest in the case of the combination of multi-frame-based action recognition and local+global cropped images ($47.80$).

Single-frame-based Action Recognition:
Let us explain the single-frame-based action recognition.
The accuracy in walking, standing, sitting, carrying, pushing/pulling, and reading is high compared to other actions.
In such actions, the mAP shows improvement (MCF vs. MHT-PAF) as shown in Table\ \ref{tb:multi}.
In order to improve the mAP, it is important to improve the accuracy of single-frame-based action recognition.

Multi-frame-based Action Recognition:
For the local cropped image, the accuracy in the case of multi-frame-based recognition is higher than that of single-frame-based recognition (single frame: $42.88$ vs. multi frames: $45.09$).
For the local+global cropped image, the accuracy in the case of multi-frame-based recognition is higher than that of single-frame-based recognition (single frame: $45.94$ vs. multi frames: $47.80$).
Therefore, the multi-frame-based action recognition is effective.
In order to leverage the multi-frame-based action recognition more effectively,
the improvement in the accuracy of human tracking is needed.

\subsection{Examples of Human Tracking and Action Recognition}
\begin{figure*}[t]
 \begin{center}
 \includegraphics[width=0.99\linewidth]{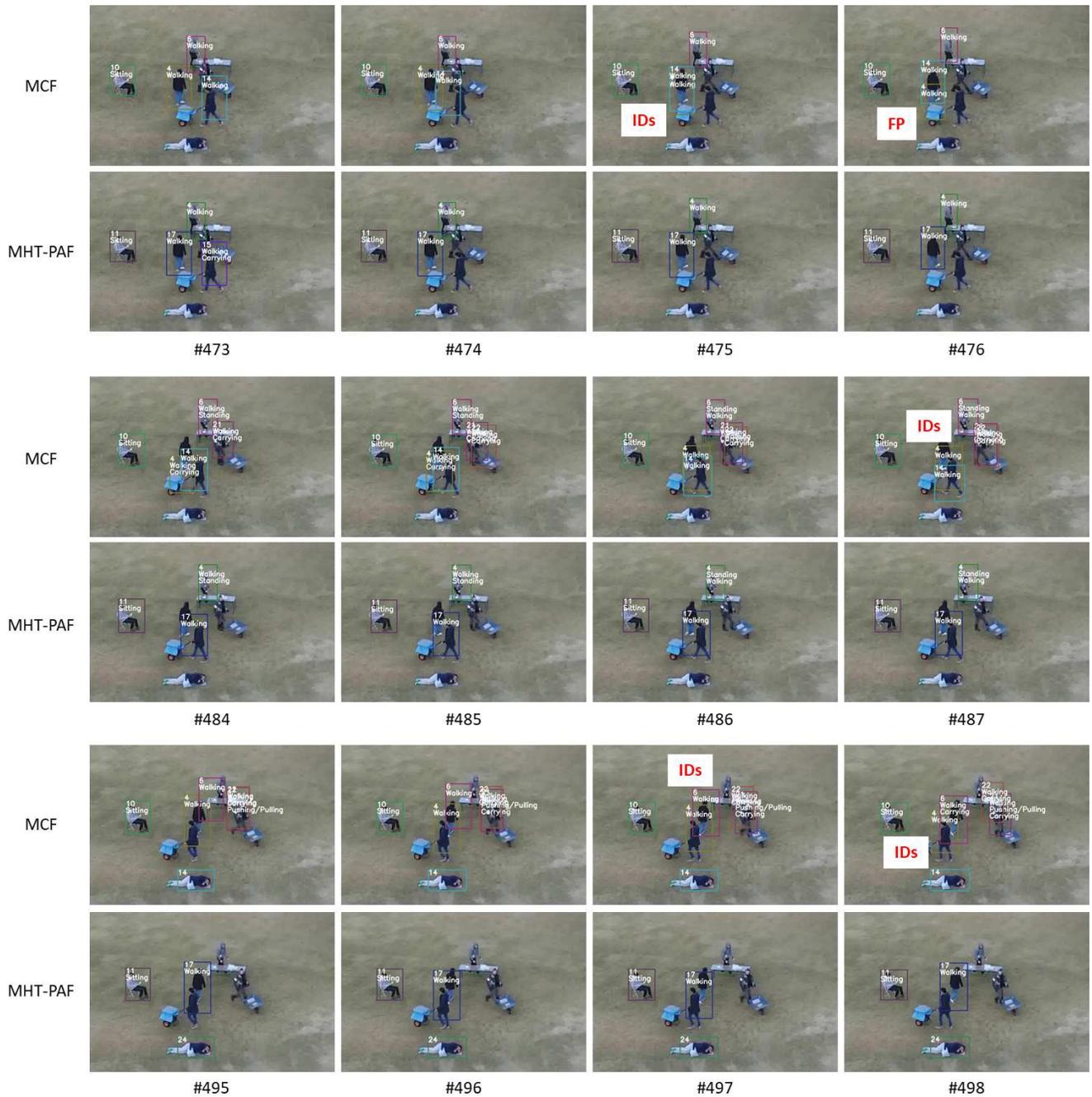}
 \end{center}
 \caption{Examples of human tracking and action recognition.}
 \label{fig:examples}
\end{figure*}

Fig.\ \ref{fig:examples} shows examples of human tracking and action recognition.
For each bounding box, estimated human ID and actions are indicated.
If the action recognition result is ``Unknown'', it is not indicated in the image.
\#(number) denotes a frame ID.
%
Fig.\ \ref{fig:examples} shows examples of video 1.2.10.
In MCF, ID switches (IDs) occur frequently (frame 475, 487, 497, and 498) and false positive (FP) occurs (frame 476).
In MHT-PAF, these ID switches and false positive are prevented.
PAF employs rich information on human action, and can avoid the miss of data association.

\section{Conclusion}
In this paper, we proposed a Multiple Human Tracking method using multi-cues including Primitive Action Features (MHT-PAF).
PAF employs a global context, rich information from multi-label actions, and a middle level feature.
Accurate human tracking using PAF can be applied to the multi-frame-based action recognition.
In the experiments, we evaluated the proposed method using Okutama-Action dataset, which consists of aerial view videos.
We verified that the human tracking accuracy (MOTA) improved $2.23$.
The number of ID switches decreased $110$ and the precision improved $2.51$ with retention of the recall.
Due to the improvement in accuracy of human tracking, the action detection accuracy (mAP) improved $0.57$.
Also, we discussed the global cropped image and single/multi-frame-based action recognition.
In the future, we will research the cooperative method that human tracking and action recognition work complementarily.

{\small
\bibliographystyle{IEEEtran}
\bibliography{egbib}
}

\end{document}